\documentclass{article}
\usepackage{iclr2015,times}
\usepackage{hyperref}
\usepackage{natbib}
\usepackage{url}
\usepackage{amsmath}
\usepackage{amssymb}
\usepackage[plain]{algorithm}
\usepackage{algorithmic}
\usepackage{wrapfig}

\usepackage{graphicx}
\usepackage{subfigure} 
\usepackage{hyperref}

\usepackage{enumitem}

\usepackage[rightcaption]{sidecap}

\title{Scheduled denoising autoencoders}

\author{
Krzysztof J. Geras\\
School of Informatics\\
University of Edinburgh\\
\texttt{k.j.geras@sms.ed.ac.uk}\\
\And
Charles Sutton\\
School of Informatics\\
University of Edinburgh\\
\texttt{csutton@inf.ed.ac.uk}\\
}

\newcommand{\xb}{\mathbf{x}}
\newcommand{\xt}{\tilde{x}}
\newcommand{\xtilde}{\tilde{\mathbf{x}}}

\iclrconference
\iclrfinalcopy

\begin{document}

\maketitle

\begin{abstract}
We present a representation learning method that learns features
at multiple different levels of scale. Working within the unsupervised
framework of denoising autoencoders, we observe that when the input
is heavily corrupted during training, the network tends
to learn coarse-grained features, whereas when the input
is only slightly corrupted, 
the network tends to learn fine-grained features.
This motivates the \emph{scheduled denoising autoencoder}, 
which starts with a high level of noise that lowers as training progresses.
We find that the resulting representation yields a significant boost on a later
supervised task compared to the original input, or to a standard denoising
autoencoder trained at a single noise level. After supervised fine-tuning our best model achieves the lowest ever reported error on the CIFAR-10 data set among permutation-invariant methods.
\end{abstract} 

\section{Introduction}

In most applications of representation learning, we wish to learn features at different levels of scale.
For example, in image data, some edges will span only a few pixels, whereas others,
such as a boundary between foreground and background, will span a large portion of the image.
Similarly, in text data, some features in the representation might model specialized topics that use only a few words. 
For example a topic about electronics would often use words such as 
``big'', ``screen'' and ``tv''. Other features model more general topics that use many different words.
Good representations should model both of these phenomena,
containing features at different levels of granularity.

Denoising autoencoders \citep{extracting, stacked, deep_sparse} provide a particularly natural
framework in which to formalise this intuition. In a denoising autoencoder,
the network is trained so as to be able to reconstruct
each data point from a corrupted version. The noise process used to perform
the corruption is chosen by the modeller, and is 
an important tuning parameter that affects the final representation.
On a digit recognition task, \cite{stacked} noticed that using a low level of noise leads to learning blob detectors, while increasing it results in obtaining detectors of strokes or parts of digits. They also recognise that either too low or too high level of noise harms the representation learnt. The relationship between the level of noise and spatial extent of the filters was also noticed by \cite{noise_level} for a different feature learning model.
 Despite impressive practical results with denoising
autoencoders, e.g. \cite{sentiment}, \cite{challenge}, the choice
of noise distribution is a tuning parameter whose effects
are not fully understood.

In this paper, we introduce \emph{scheduled denoising autoencoders} (ScheDA),
which are based on the intuition that by training the same network at multiple noise levels,
we can encourage it to learn features at different scales. 
The network is trained with a schedule of gradually
decreasing noise levels. At the initial, high noise levels, the training data 
is highly corrupted, which forces the network to learn more global, coarse-grained
features of the data. At low noise levels, the network is able
 to learn features for reconstructing finer details of the training data.
At the end of the schedule, the network will include a combination of
both coarse-grained and fine-grained features.

This idea is reminiscent of continuation methods, which have also been applied to neural networks \citep{curriculum_learning}. The motivation of this work is significantly different though. 
Our goal is to encourage
the network to learn a more diverse set of features,
 some which are similar to features learnt
at the initial noise level, and others which are similar to features
learnt at the final noise level. In Section \ref{sec:similarity_experiment}, we verify quantitatively
that this happens.

Experimentally, we find on both image and text data that scheduled denoising
autoencoders learn better representations than standard denoising
autoencoders,
as measured by the features' performance on a supervised task. 
On both classification tasks, the representation from ScheDA yields lower test error than that from a denoising autoencoder
trained at the best single noise level. After supervised fine-tuning our best ScheDA model achieves the lowest ever reported error on the CIFAR-10 data set among permutation-invariant methods.

\section{Background}
The core idea of learning a representation by learning to reconstruct artificially corrupted training data dates back at least to the work of \citet{continuous_attractors}, who suggested using a recurrent neural network for this purpose. Using unsupervised layer-wise learning of representations for classification purposes appeared later in the work of \citet{greedy_layer} and \citet{fast_algorithm}.

The denoising autoencoder (DA) \citep{extracting} is based on the same intuition as the work of \citet{continuous_attractors} that that
a good representation should contain enough information to reconstruct 
corrupted versions of an original input.
Let $\mathbf{x} \in \mathbb{R}^d$ be the input to the network.
The output of the network is a hidden representation $\mathbf{y} \in \mathbb{R}^{d'}$,
which is simply computed as 
$f_{\theta}(\mathbf{x}) = s(\mathbf{W}\mathbf{x} + \mathbf{b})$,
where the matrix $\mathbf{W} \in \mathbb{R}^{d' \times d}$ and the vector
$\mathbf{b} \in \mathbb{R}^{d'}$ are the parameters of the network,
and $s$ is a typically nonlinear transfer function, such as a sigmoid.
We write $\theta = (\mathbf{W}, \mathbf{b})$.
The function $f$ is called an \emph{encoder}
because it maps the input to a hidden representation. In an autoencoder, we have also a \emph{decoder} that ``reconstructs'' the input
vector from the hidden representation, which is used when training the network.
The decoder has a similar form to the encoder, namely,
$g_{\theta'}(\mathbf{y}) = t(\mathbf{W'}\mathbf{y} + \mathbf{b'})$, except
that here $\mathbf{W'} \in \mathbb{R}^{d \times d'}$ and $\mathbf{b'} \in \mathbb{R}^{d}$. It can be useful to allow the transfer function $t$ for the decoder
to be different from that for the encoder. 
Typically, $\mathbf{W}$ and $\mathbf{W}'$ are constrained by $\mathbf{W}' = \mathbf{W}^T$ by analogy to the interpretation of principal components analysis as a linear encoder and decoder.

During training, our objective is to learn the encoder parameters $\mathbf{W}$ and $\mathbf{b}$.
As a byproduct, we will need to learn the decoder parameters $\mathbf{b'}$ as well.
We do this by defining a \emph{noise distribution} $p(\xtilde | \xb, \nu)$. The amount of corruption is controlled by
a parameter $\nu$. We train the autoencoder weights to be able to reconstruct
a random input from the training distribution $\xb$ from its corrupted version $\xtilde$ by running the encoder and the decoder in sequence. Formally, this process is described by the objective function
\begin{equation}
{\theta}^*, {\theta'}^{*} = \operatorname*{arg\,min}_{\theta, \theta'} \mathbb{E}_{(X, \tilde{X})} \left[ L\left(X, g_{\theta'}(f_{\theta}(\tilde{X}))\right) \right],
\label{eq:objective}
\end{equation}
where $L$ is a loss function over the input space, such as squared error. Typically we minimize this objective function using stochastic gradient descent with mini-batches, where at each iteration
we sample new values for both the uncorrupted and corrupted inputs.

In the absence of noise, this model is known simply as an autoencoder or autoassociator. A classic result \citep{ae_pca} states that when $d' < d$, then under certain conditions, an autoencoder learns the same subspace as PCA. If the dimensionality of the hidden representation is too large, i.e., if $d' > d$, 
then the autoencoder can obtain zero reconstruction error simply by learning the identity map.
In a denoising autoencoder, in contrast, the noise forces the model to learn interesting structure even when there are a large 
number of hidden units. Indeed, in practical denoising autoencoders, often the best results are found with \emph{overcomplete} representations for which $d' > d$.

There are several tuning parameters here, including the noise distribution, the transformations $s$ and $t$ and the loss function $L$. 
For the loss function $L$, for continuous $\xb$, squared error can be used.
For binary $\mathbf{x}$ or $\mathbf{x} \in [0,1]$, as we consider in this paper, it is common to use the \emph{cross entropy} loss,
\begin{equation*}
L(\mathbf{x}, \mathbf{z}) = -\sum_{i=1}^{D} \left(x_i\log{z_i} + (1-x_i)\log{(1-z_i)}\right).
\label{eq:cross_entropy}
\end{equation*}
For the transfer functions, common choices include the sigmoid
$s(v) = \frac{1}{1+e^{-v}}$ for both the encoder and decoder, or to use a rectifier $s(v) = \mathrm{max}(0, v)$
in the encoder paired with sigmoid decoder.

One of the most important parameters in a denoising autoencoder is 
the noise distribution $p$. For continuous $\xb$, Gaussian noise $p(\xtilde | \xb, \nu) = N(\xtilde; \xb, \nu)$ can be used.
For binary $\mathbf{x}$ or $\mathbf{x} \in [0,1]$, it is most common to use \emph{masking noise}, that is,
for each $i \in 1, 2, \ldots d$, we sample $\xt_i$ independently as
\begin{equation}
p(\xt_i | x_i, \nu) = 
\begin{cases}
0 & \text{with probability $\nu$}, \\
x_i & \text{otherwise}.
\end{cases}
\end{equation}
In either case, the level of noise $\nu$ affects the degree of corruption of
the input. If $\nu$ is high, the inputs are more heavily corrupted
during training. The noise level has a significant effect on the representations learnt. 
For example, if the input data are images, masking only a few pixels will bias the process of learning the representation to deal well with local corruptions. On the other hand, masking very many pixels will push the algorithm to use information from more distant regions.

It is also possible to train multiple layers of representations with denoising autoencoders by training
a denoising autoencoder with data mapped to a representation learnt by another denoising autoencoder. This model is known as the stacked denoising autoencoder \citep{extracting, stacked}.

\section{Scheduled denoising autoencoders}

Our goal is to learn a single representation
that combines the best aspects of representations learnt at different levels of noise.
The scheduled denoising autoencoder (ScheDA) aims to do this by training
a single DA sequentially using a \emph{schedule} of noise levels, such that
$\nu_0 > \cdots > \nu_T \geq 0$. The initial noise level $\nu_0$ is chosen to be a high
noise level that corrupts most of the input.
The final noise level $\nu_T$ is chosen to be lower than the optimal noise level for a standard
DA, i.e., chosen via a held-out validation set or by cross-validation. In pseudocode,
\begin{algorithmic}
\WHILE{$\theta$ not converged}
  \STATE Take a stochastic gradient step on \eqref{eq:objective}, using noise level $\nu_0$.
\ENDWHILE
\FOR{$t$ in $1, \ldots, T$}
  \STATE$\nu_t := \nu_{t-1} - \Delta\nu$
  \FOR{$K$ steps}
    \STATE Take a stochastic gradient step on \eqref{eq:objective}, using noise level $\nu_t$.
  \ENDFOR
\ENDFOR
\end{algorithmic}
This method is reminiscent of deterministic annealing \citep{rose98},
which has been applied to clustering problems, in which a sequence
of clustering problems are solved at a gradually lowered noise level. 
However, the meaning of ``noise level'' is very different. In deterministic
annealing, the noise is added to the mapping between inputs and cluster labels.
This is to encourage data points to move between cluster centroids early in 
the optimization process. 

\begin{figure*}[tb!]
\begin{tabular}{cc}
\includegraphics[scale=0.485, trim=40mm 108mm 30mm 108mm]{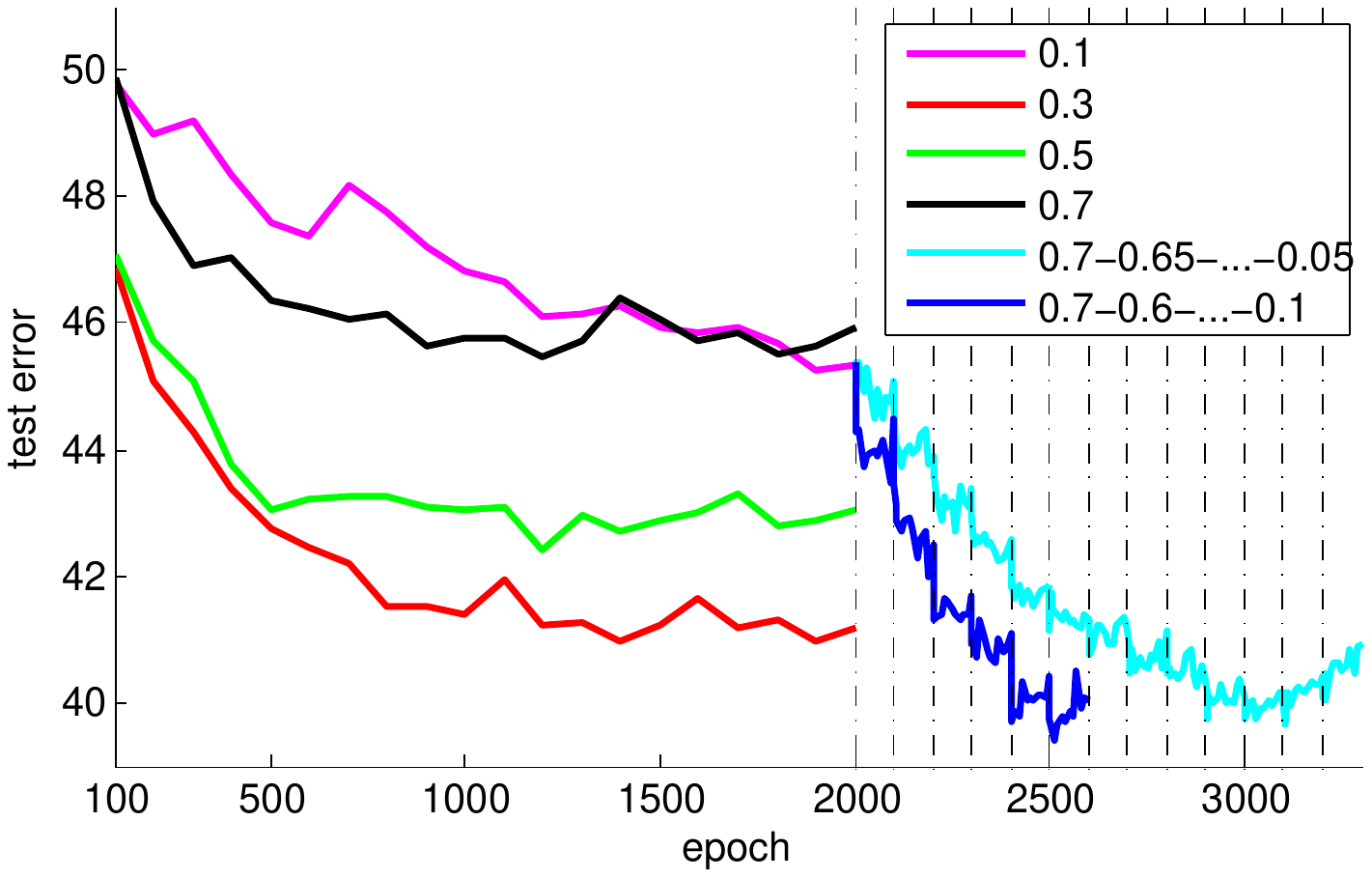} &
\includegraphics[scale=0.485, trim=35mm 108mm 30mm 108mm]{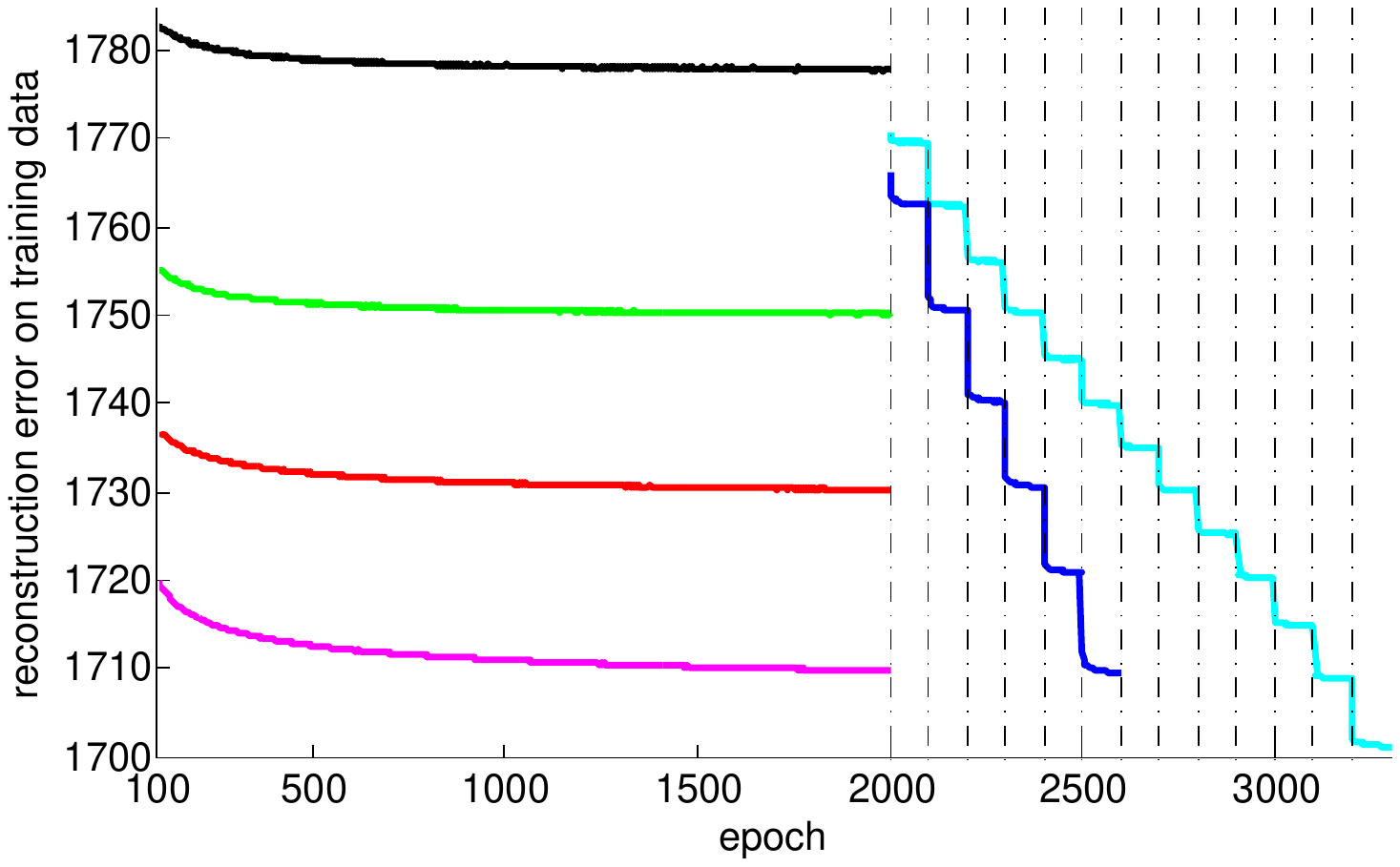}\\
\end{tabular}
\caption{Experimental results with CIFAR-10 for 2000 hidden units. Test errors (left) and reconstruction errors on training set (right) as a function of the number of epochs. Dashed lines indicate a point when the level of noise was changed. Test errors were measured every 100 training epochs initially (during the first 2000 epochs). After each change of the noise level, test error was measured after the first, the third, the fifth epoch and then after every ten epochs. Reconstruction errors were measured after each training epoch for data corrupted with the noise level used for training at that epoch. For clarity, we do not show the results of DA (0.2), DA (0.4) and DA (0.6), which yield higher test error than DA (0.3).}
\label{fig:CIFAR10_errors_and_reconstructions}
\end{figure*}

ScheDA is also conceptually related to curriculum learning \citep{curriculum_learning}
and continuation methods more generally \citep{continuation:book}.
In curriculum learning, the network is trained on a sequence of learning problems that have the
property that the earlier tasks are ``easier'' than later tasks. In ScheDA, it is less obvious that the earlier tasks are easier since the lowest achievable reconstruction error is actually higher at the earlier high noise levels than at the later low noise levels. We observe this 
in practice (cf. \autoref{fig:CIFAR10_errors_and_reconstructions}). On the other hand, we found that, for a given learning rate, the reconstruction error converges to a local minimum faster with large $\nu$'s (cf. the right panel of \autoref{fig:CIFAR10_errors_and_reconstructions}). Thus, even though the problems that ScheDA starts with are harder in absolute terms, finding the local minima for these problems is easier. This can be understood given the insight provided by the work of \cite{score_matching}, who has shown that, for a DA trained with Gaussian noise and squared error, minimising reconstruction error is equivalent to matching the score (with respect to the input) of a nonparametric Parzen density estimator of the data, which depends on the level of noise. An implication of this viewpoint is that if the density learnt by the Parzen density estimator is harder to represent, it makes the DA learning problem harder too. Convolving the data with a high level of noise transforms the data generating distribution into a much smoother one, which is easier to capture. As the noise level is reduced, the density becomes more multimodal and harder to represent.

\section{Experiments}
\label{sec:experiments}

\begin{table}[tb!]
\centering
\caption{Test errors on CIFAR-10 data set. Each ScheDA is characterised by the sequence of noise levels it was trained with and the number of epochs for which it was trained at each level of noise after the first noise level switch. Each row shows the best DA and the best ScheDA for a given number of hidden units, choosing the learning rate, the number of training epochs and the noise level using the error on the validation set.}
\label{tab:best_CIFAR10}
\vspace{5pt}
\begin{tabular}{| c | c | c | c | c | c | c |}         
  \cline{1-1} \cline{3-4} \cline{6-7}
  \textbf{hidden units} & & \textbf{best DA} & \textbf{test error} & & \textbf{best ScheDA} & \textbf{test error}\\
  \cline{1-1} \cline{3-4} \cline{6-7}
  1000 & & 0.4 & 45.34\% & & 0.4$\to$0.3$\to$0.2, $K$=50& 43.01\%\\
  \cline{1-1} \cline{3-4} \cline{6-7}
  2000 & & 0.3 & 41.95\% & & 0.7$ \to$0.65$\to\ldots\to$0.2$\to$0.15, $K$=100& 40.1\%\\
  \cline{1-1} \cline{3-4} \cline{6-7}
  5000 & & 0.1 & 38.64\% & & 0.2$\to$0.15$\to$0.1$\to$0.05, $K$=50& 36.77\%\\
  \cline{1-1} \cline{3-4} \cline{6-7}
\end{tabular}
\end{table}

We evaluate ScheDA on two different data sets, an image classification data set (CIFAR-10), and a text classification data set (Amazon product reviews, results in the supplementary material). We use a procedure similar to one used, for example, by \cite{an_analysis}\footnote{We do not use any form of pooling, keeping our setup invariant to the permutation of the features.}. That is, in all experiments, we first learn the representation in an unsupervised fashion and then use the learnt representation within a linear classifier as a measure of its quality. In both experiments, in the unsupervised feature learning stage, we use masking noise as the corruption process, a sigmoid encoder and decoder and cross entropy loss (\autoref{eq:cross_entropy})\footnote{We also tried a rectified linear encoder combined with sigmoid decoder on the Amazon data set. The results were very similar, so we do not show them here.} following \cite{extracting, stacked}. All experiments with learning the representations were implemented using the Theano library \citep{theano}. To do optimisation, we use stochastic gradient descent with mini-batches. For the classification step, we use $L2$-regularised logistic regression implemented in LIBLINEAR \citep{liblinear}, with the regularisation parameter chosen to minimise the validation error.

\subsection{Image recognition}
\label{sec:cifar10}
We use the CIFAR-10 \citep{learning_multiple} data set for experiments with vision data. This data set consists of 60000 colour images spread evenly between ten classes. There are 50000 training and validation images and 10000 test images. Each image has a size of 32x32 pixels and each pixel has three colour channels, which are represented with a number in $\{0, \ldots, 255\}$. We divide the training and validation set into 45000 training instances and 5000 validation instances. The only preprocessing step we use is dividing the intensity of every pixel by 255 to get numbers in $[0,1]$.

To get the strongest possible DAs trained with a single noise level, we choose the noise level, learning rate and number of training epochs in order to minimise classification error on the validation set. We try all combinations of the following values of the parameters: noise level $\in \{0.7, 0.6, 0.5, 0.4, 0.3, 0.2, 0.1, 0.05\}$, learning rate $\in \{0.002, 0.01, 0.05\}$, number of training epochs $\in \{100, 200, \ldots, 2000\}$. We choose these parameters separately for each size of the hidden layer $\in \{1000, 2000, 5000\}$.

To train ScheDA models, we first pick the best DA for each level of noise we consider, optimising the learning rate and the number of training epochs with respect to the validation error. Starting from the best DA for given $\nu_0$, we continue the training, lowering the level of noise from $\nu_{t-1}$ to $\nu_{t} := \nu_{t-1} - \Delta \nu$ and training the model for $K$ epochs. We repeat this noise reduction $T$ times. In our experiments we consider $\Delta \nu \in \{0.05, 0.1\}$ and $K \in \{50, 100\}$. We use the learning rate of 0.01 for this stage as it turned out to always be optimal or very close to optimal for the standard DA\footnote{Note that tuning this parameter could only help ScheDAs and would not affect the baselines.}. We pick the combination of parameters $(\nu_0, \Delta\nu, K)$ and the number of noise reduction steps, $T$, using the validation error of a classifier after the last training epoch at each level of noise $\nu_t$. We denote a DA trained with the level of noise $\nu$ by DA ($\nu$) and ScheDA trained with a schedule of noise levels $\nu_0$, $\nu_1$, ..., $\nu_T$ by ScheDA ($\nu_0$$\to$$\nu_1$$\to$...$\to$$\nu_T$).

The error obtained by the classifier trained with raw pixel data equals 59.78\%. A summary of the performance of DAs and ScheDAs for each number of hidden units can be found in \autoref{tab:best_CIFAR10}. For each size of the hidden layer we tried, ScheDA easily outperforms DA, with a relative error reduction of about 5\%. Our best model achieves the error of 36.77\%. Interestingly, our method is very robust to the parameters $(\nu_0, \Delta\nu, K)$ of the schedule. See Section \ref{sec:schedules} for more details. Those results do not use supervised fine-tuning. Supervised fine-tuning of our best model yielded the error of 35.7\%, which, to our knowledge, is the lowest ever reported error for permutation invariant CIFAR-10, outperforming \cite{fastfood} who achieved the error of 36.9\% and \cite{zero_bias}, who achieved the error of 36.1\%. We describe the details of our supervised fine-tuning procedure in the supplementary material.

\autoref{fig:CIFAR10_errors_and_reconstructions} shows the test errors and reconstruction errors on the training data as a function of the training epoch for selected DAs and ScheDAs with 2000 hidden units.
It is worth noting that, even though the schedules exhibiting the best performance go below the optimal $\nu$ for DA, training for many iterations with a level of noise that is too low hurts performance (see the final parts of the schedules shown in \autoref{fig:CIFAR10_errors_and_reconstructions}). This may be due to the fact that structures learnt at low noise levels are too local to help generalisation.

The performance of our method does not appear to be solely due to better optimisation of the training objective. For example, DA (0.1) trained for 3000 epochs has a lower reconstruction error on the training data than the ScheDA (0.7$\to$0.6$\to$...$\to$0.1) shown in \autoref{fig:CIFAR10_errors_and_reconstructions}, while the test error it yields is higher by about 5\%.

\begin{figure*}[tb!]
\centering
\begin{minipage}{1.0\textwidth}
\begin{tabular}{cccc}
	\hspace{-0.22cm}
	\includegraphics[scale=0.55, trim=0mm 0mm 0mm 0mm, clip=true]{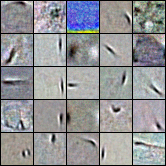}&
	\hspace{-0.22cm}
	\includegraphics[scale=0.55, trim=0mm 0mm 0mm 0mm, clip=true]{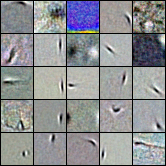}&
	\hspace{-0.22cm}
	\includegraphics[scale=0.55, trim=0mm 0mm 0mm 0mm, clip=true]{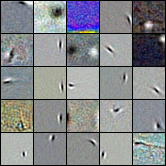}&
	\hspace{-0.22cm}
	\includegraphics[scale=0.55, trim=0mm 0mm 0mm 0mm, clip=true]{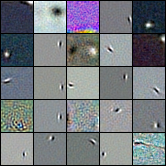}\\
	\scriptsize{DA (0.7)} & \scriptsize{ScheDA (0.7$\to$0.6$\to$0.5)} & \scriptsize{ScheDA (0.7$\to$0.6$\to$...$\to$0.3)} & \scriptsize{ScheDA (0.7$\to$0.6$\to$...$\to$0.1)} 
\end{tabular}
\vspace{-8pt}
\caption{A sample of filters (rows of the matrix $\mathbf{W}$) learnt from CIFAR-10 with DA (0.7) and ScheDAs starting with $\nu_0 = 0.7$. All sets of filters are similar, but those that were post-trained with low level of noise are sharper. With schedules that end at a lower level of noise, the filters become more local but not as much as when only training with a low level of noise (cf. \autoref{fig:CIFAR10_filters_low_high}).}
\label{fig:CIFAR10_filters}
\end{minipage}
\end{figure*}
 
\begin{figure*}[tb!]
\begin{minipage}{1.0\textwidth}
\centering
\begin{tabular}{cc}
	\hspace{-0.22cm}
	\includegraphics[scale=0.64, trim=0mm 0mm 0mm 0mm, clip=true]{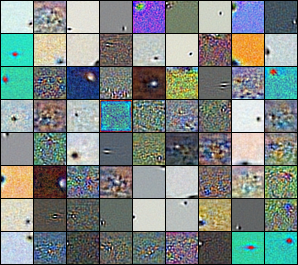}&
	\hspace{-0.22cm}
	\includegraphics[scale=0.64, trim=0mm 0mm 0mm 0mm, clip=true]{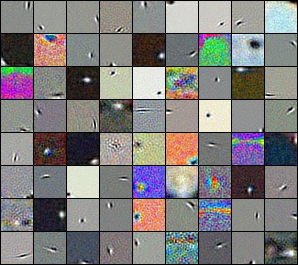}\\
	\tiny{DA (0.1)} & \tiny{ScheDA (0.7$\to$0.6$\to$0.5$\to$0.4$\to$0.3$\to$0.2$\to$0.1)}
\end{tabular}
\vspace{-8pt}
\caption{Samples of filters (rows of the matrix $\mathbf{W}$) learnt from CIFAR-10 with a low fixed noise level (left) and filters learnt with an initially high level of noise and post-trained with a schedule of lower levels of noise (right). These two sets of filters are visually very different. There are fewer edge detector filters among these learnt only with a low level of noise and those that are edge detectors are more local.}
\label{fig:CIFAR10_filters_low_high}
\end{minipage}
\end{figure*}

\begin{figure*}[tb!]
\begin{minipage}{1.0\textwidth}
\centering
\vspace{-7pt}
\begin{tabular}{ccc}
	\hspace{-0.4cm}
	\includegraphics[scale=0.58, trim=62mm 120mm 65mm 120mm, clip=true]{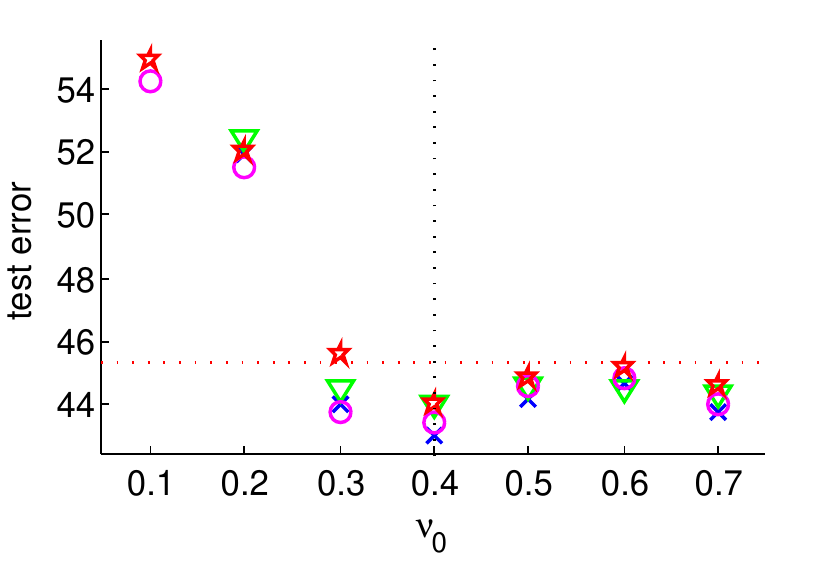}&
	\hspace{-0.4cm}
	\includegraphics[scale=0.58, trim=66mm 120mm 65mm 120mm, clip=true]{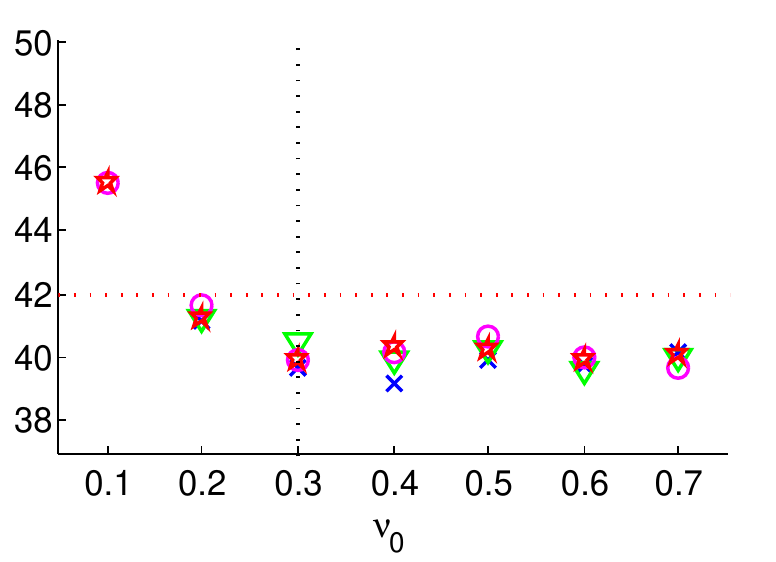}&
	\hspace{-0.4cm}
	\includegraphics[scale=0.58, trim=66mm 120mm 65mm 120mm, clip=true]{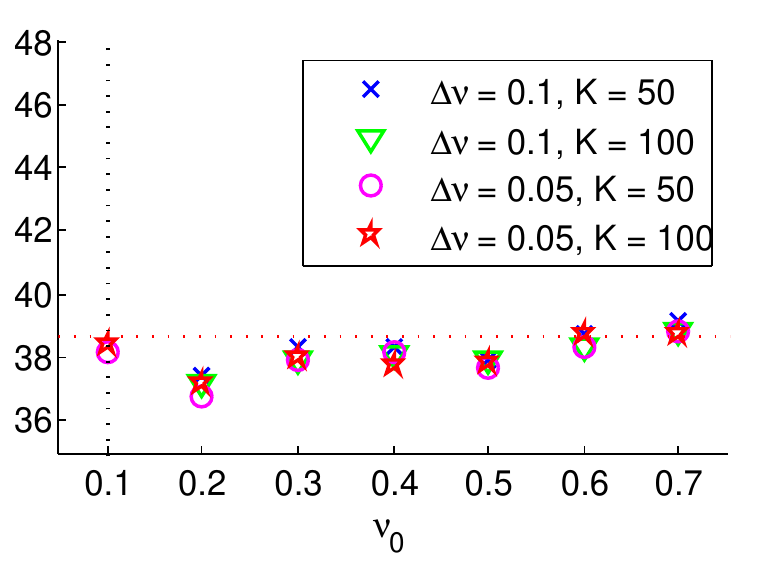}\\
	1000 hidden units & 2000 hidden units & 5000 hidden units
\end{tabular}
\vspace{-8pt}
\caption{Comparison of different schedules for ScheDA. For each size of hidden layer, the vertical line indicates the optimal level of noise for a DA, i.e., the level of noise which allowed to train representation yielding the lowest error on the validation set, while the horizontal line indicates the test error obtained with this representation.}
\label{fig:CIFAR10_schedules_all}
\end{minipage}
\end{figure*}

The features learnt with ScheDA are visibly noticeably different from those learnt with any single level of noise as they contain a mixture of features that could be found for various values of $\nu$. \autoref{fig:CIFAR10_filters} and \autoref{fig:CIFAR10_filters_low_high} display visualisations of the filters learnt by a DA and ScheDA. It can be seen in \autoref{fig:CIFAR10_filters} that when training ScheDA the features across the consecutive levels of noise are similar, which indicates that it is the initial training with a higher level of noise that puts the optimisation procedure in a basin of attraction of a different local minimum, which would not be otherwise achievable. This is shown in \autoref{fig:CIFAR10_filters_low_high}, which visualises features learnt by a DA trained only with a low noise level, DA (0.1), and those learnt with ScheDA (0.7$\to$0.6$\to$...$\to$0.1). The set of features learnt by DA (0.1) contains more noisy features and very few edge detectors, which are all very local. In contrast, features learnt with the schedule contain a more diverse set of edge detectors which can be learnt with high noise level (\autoref{fig:CIFAR10_filters}) as well as some blob detectors which can be learnt with a low noise level (\autoref{fig:CIFAR10_filters_low_high}).

\subsubsection{Robustness to the choice of schedule}
\label{sec:schedules}
Our method is very robust to the choice of the parameters of the schedule, $\nu_0$, $\Delta\nu$ and $K$. \autoref{fig:CIFAR10_schedules_all} shows the performance of ScheDA for different values of those parameters. For 1000 and 2000 hidden units for all schedules ScheDA performed better than the best DA, as long as the initial level of noise $\nu_0$ was not lower than the level of noise yielding the best DA. For 5000 hidden units, ScheDA also performed better than DA, except for the model trained with $\nu_0 = 0.7$. These results suggest than ScheDA's performance is superior to DA as long as the initial level of noise is not too large and not below the optimal level of noise for DA.

We also examined whether it is necessary for the schedule to be decreasing. To investigate this, we trained ScheDAs using the same procedure as before, except that the levels of noise were increasing. The networks had 2000 hidden units and they started with the best DA (over the learning rate and the number of training epochs) for each $\nu_0$. We considered $\nu_0 \in \{0.1, 0.2, 0.3, 0.4\}$, $K \in \{50, 100\}$ and $\Delta\nu \in \{0.05, 0.1\}$. The largest possible final noise level $\nu_T$ was 0.7. To evaluate all combinations of these hyperparameters ($\nu_0$, $\nu_T$, $\Delta\nu$ and $K$) we used the validation set. We set the learning rate to 0.01 as it worked optimally or very close to optimally in all previous experiments and we also used this value in the experiments with decreasing schedules. The best model we obtained this way used the schedule 0.3$\to$0.35 and $K=100$. Its test error was 41.97\%, just a little worse than a DA trained with $\nu=0.3$ (achieving the test error of 41.95\%). For comparison, ScheDA (0.1$\to$0.2$\to$0.3) with $K=100$ yielded the test error of 44.99\% and ScheDA (0.3$\to$0.4$\to$0.5$\to$0.6$\to$0.7) with $K=100$ yielded the test error of 46.7\%. These results provide some evidence that the initial noise levels puts the optimisation procedure in a basin of attraction of a local minimum that can be favourable, as we observe for ScheDA when starting training with higher noise levels, or detrimental, as we see here.

\subsubsection{Concatenating sets of features learnt with different noise levels}
\label{sec:feature-concat}

\begin{wrapfigure}{right}{6.5cm}
\vspace{-0pt}
\begin{center}
\includegraphics[scale=0.5, trim=70mm 95mm 70mm 113mm]{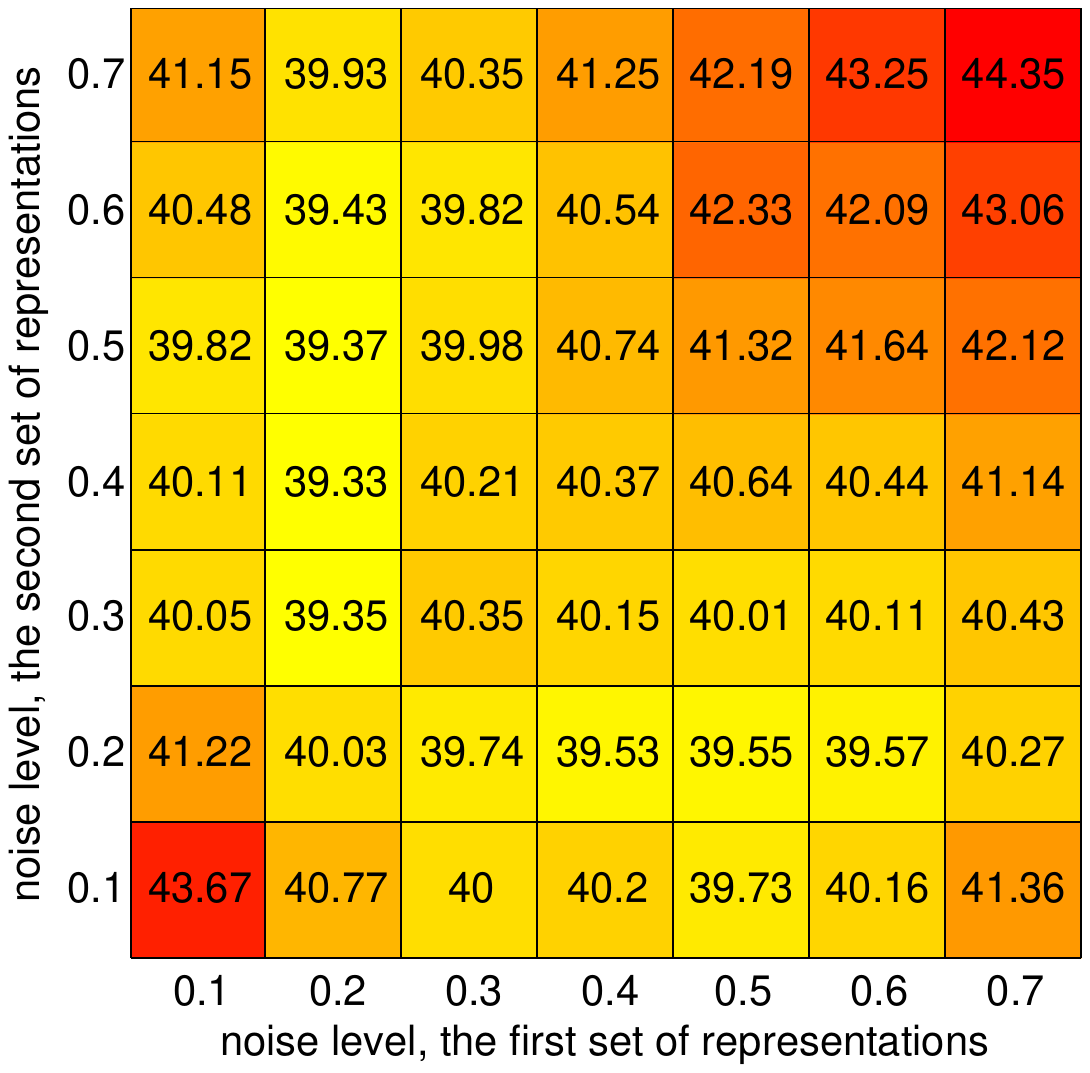}
\end{center}
\vspace{-7pt}
\caption{Test errors yielded by representations constructed by concatenating representations learnt with various levels of noise.
This allows representations that are otherwise weak separately to achieve low test errors, e.g. for $\nu=0.1$ and $\nu=0.5$ (cf. \autoref{fig:CIFAR10_errors_and_reconstructions}).}
\vspace{-40pt}
\label{fig:CIFAR10_compositions}
\end{wrapfigure}

To explain the results above, we examine whether features learnt with different noise levels contain different information about the data. To explore this, we trained two sets of representations with 2000 hidden units independently with a standard DA. DAs in the first set were initialised with a randomly drawn set of parameters $\theta_1$ and DAs in the second set were initialised with a different randomly drawn set of parameters ${\theta}_2$. Each set contained representations learnt with $\nu=0.1$, $\nu=0.2$, ..., $\nu=0.7$. Then we gathered all 49 possible pairs of representations between the two sets and concatenated representations within each pair, creating representations with 4000 features. The errors yielded by classifiers using these representations can be found in \autoref{fig:CIFAR10_compositions}. The important observation here is that, even though concatenating two representations learnt with the same $\nu$ but with different initialisations results in a better representation (cf. \autoref{fig:CIFAR10_errors_and_reconstructions}), concatenating representations with different $\nu$'s yields even lower errors.

This is another piece of evidence strengthening our hypothesis that having both local and global features in the representation learnt with ScheDA helps classification. Note, however, that even though concatenating representations
learnt with different $\nu$ helps, ScheDA is clearly a better model. For a fair comparison, we trained ScheDA with 4000 units using, which matches the number of hidden units in the concatenated architecture. While the best concatenated DA achieved 39.33\% (cf. \autoref{fig:CIFAR10_compositions}), the best ScheDA achieved 37.77\%.


\subsubsection{Comparing sets of features}
\label{sec:similarity_experiment}

Having confirmed that using both features learnt with different levels of noise indeed helps classification, we experimentally verify the hypothesis that the final representation trained with ScheDA (0.7$\to$0.6$\to$...$\to$0.1) contains both features similar to those learnt with low levels of noise (local features) and high levels of noise (global features).

Intuitively, two features are similar if they are active for the same set of inputs. We define the activation vector $\mathbf{a}_i$ for feature $i$ as the vector
containing the activation of the feature over all the data points.
More formally, if $\mathbf{w}_i$ is the weight vector for feature $i$, $b_i$ is the bias for feature $i$ and 
$\mathbf{x}_n$ is a data item, the activation vector is 
$\mathbf{a}_i = [a_{i1}, \ldots, a_{iN}]$, where 
$a_{in} = \mathrm{sigmoid}(\mathbf{w}_i\mathbf{x}_n + b_i)$. Here $N$ is the total number of data items, the total number of features is $I$. 

We compute the activation vector for all features from eight different 
autoencoders: DA (0.1), DA (0.2), ..., DA (0.7) and ScheDA (0.7$\to$0.6$\to$...$\to$0.1).
We denote the resulting activation vectors $\mathbf{a}^{0.1}_i$, ..., $\mathbf{a}^{0.7}_i$ and $\mathbf{a}^{S}_i$, respectively. Now for each feature in ScheDA (0.7$\to$0.6$\to$...$\to$0.1) we can find the closest feature among those learnt with DA (0.1), DA (0.2), ..., DA (0.7). To do this,
we compute cosine similarities $\mathrm{cos}(\mathbf{a}^{S}_i, \mathbf{a}^{0.1}_j)$, $\mathrm{cos}(\mathbf{a}^{S}_i, \mathbf{a}^{0.2}_j)$, ..., $\mathrm{cos}(\mathbf{a}^{S}_i, \mathbf{a}^{0.7}_j)$ for all pairs $(i, j)$. Finally, we compute
$\mathrm{C}_{0.1}$, the number of ScheDA features that are closest to a feature
from DA (0.1) as $\mathrm{C}_{0.1} = \sum_{i=1}^{I} 1[\mathrm{max}_j\mathrm{cos}(\mathbf{a}^{S}_i, \mathbf{a}^{0.1}_j) > \{\mathrm{max}_j\mathrm{cos}(\mathbf{a}^{S}_i, \mathbf{a}^{0.2}_j), \mathrm{max}_j\mathrm{cos}(\mathbf{a}^{S}_i, \mathbf{a}^{0.3}_j), ..., \mathrm{max}_j\mathrm{cos}(\mathbf{a}^{S}_i, \mathbf{a}^{0.7}_j)\}]$ and similarly for $\mathrm{C}_{0.2}$, $\mathrm{C}_{0.3}$, ..., $\mathrm{C}_{0.7}$.
To see how much ScheDA differs in that respect from the standard DA trained only at the final level of noise for ScheDA, we also performed the same procedure as described above, but comparing to features learnt by DA* (0.1), which is the same as DA (0.1) but starting from a different random initialisation. We found that ScheDA contains more features similar to those learnt with higher noise levels than DA* (0.1) (see \autoref{tab:comparing}). This confirms our expectation that the ScheDA representation retains a large number of more global features from the earlier noise levels. We also put the same numbers for DA* (0.7) for comparison.

\begin{table}
\centering
\caption{Comparison of features of ScheDA and DA. The first row shows how many ScheDA (0.7$\to$0.6$\to$...$\to$0.1) features, out of 2000 in total, were closest to a feature learnt by DA (0.1), ..., DA (0.7). It demonstrates that ScheDA combines information that would be learnt from DAs at varying noise levels. The second and third row are baselines for comparison (see text for details).}
\vspace{5pt}
\begin{tabular}{ c | c | c | c | c | c | c | c |}      
  \cline{2-8}   
  & \small{\textbf{DA (0.1)}} & \small{\textbf{DA (0.2)}} & \small{\textbf{DA (0.3)}} & \small{\textbf{DA (0.4)}} & \small{\textbf{DA (0.5)}} & \small{\textbf{DA (0.6)}} & \small{\textbf{DA (0.7)}}\\
  \cline{1-1} \cline{2-8}
  \multicolumn{1}{|c|}{\small{\textbf{ScheDA}}} & \small{374} & \small{550} & \small{444} & \small{299} & \small{169} & \small{92} & \small{72}\\
  \cline{1-1} \cline{2-8}
  \multicolumn{1}{|c|}{\small{\textbf{DA* (0.1)}}} & \small{1247} & \small{465} & \small{167} & \small{54} & \small{21} & \small{12} & \small{7} \\
  \cline{1-1} \cline{2-8}
  \multicolumn{1}{|c|}{\small{\textbf{DA* (0.7)}}} & \small{25} & \small{30} & \small{72} & \small{165} & \small{308} & \small{587} & \small{813} \\
  \cline{1-1} \cline{2-8}
\end{tabular}
\label{tab:comparing}
\end{table}

\section{Composite denoising autoencoder}

The observation that more diverse representations lead to a better discriminative performance can be exploited more explicitly than in ScheDA. Instead of training all of the hidden units with a sequence of noise levels, we can partition the
hidden units, training each subset of units with a different noise level. This can be done by defining the hidden representation and the reconstruction to be

$\mathbf{y} = \left[f\left(\tilde{\mathbf{x}}_{\nu_1}\mathbf{W}_1 + \mathbf{b}_1 \right), \ldots, f\left(\tilde{\mathbf{x}}_{\nu_S}\mathbf{W}_S + \mathbf{b}_S \right)\right]$ and\\ 
$\mathbf{z} = g\left( \sum_{s=1}^{S} f\left( \tilde{\mathbf{x}}_{\nu_s}\mathbf{W}_s + \mathbf{b}_s \right) \mathbf{W}^T_s + \mathbf{b'} \right)$,

\begin{wrapfigure}{right}{6cm}
\centering
\vspace{-30pt}
\includegraphics[scale=0.5, trim=60mm 110mm 60mm 120mm]{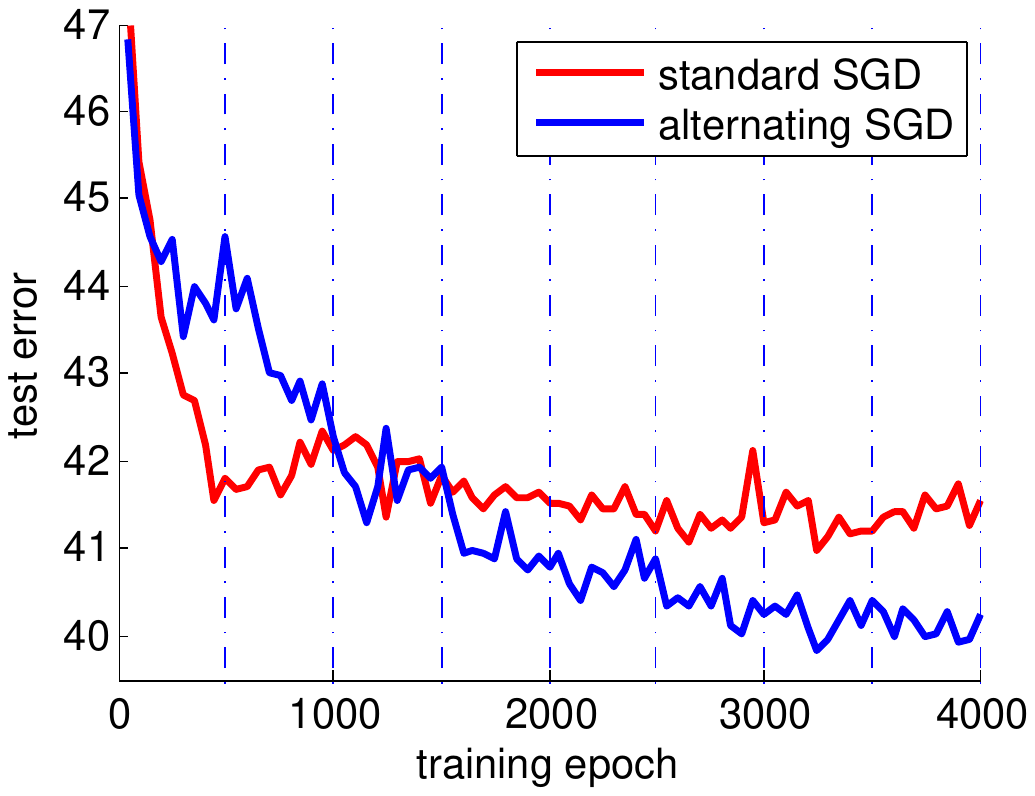}
\vspace{-5pt}
\caption{Test errors for a composite denoising autoencoder using two levels of noise, $\nu=0.2$ and $\nu=0.4$, and with 2000 hidden units divided equally between the two levels of noise. Dashed lines indicate the epochs when optimisation switched between updating different sets of parameters.}
\label{fig:composite}
\vspace{-35pt}
\end{wrapfigure}

where $\tilde{\mathbf{x}}_{\nu_s}$ denotes an input $\mathbf{x}$ corrupted with the level of noise $\nu_s$. We call this a composite DA. Our preliminary experiments show that, even when using only two noise levels, it outperforms a standard DA and performs on par with ScheDA. Successful learning of the parameters is more complicated though. We found that standard SGD (updating all parameters at each epoch) performs much worse than a version of the SGD alternating between updating parameters associated with the two levels of noise. See \autoref{fig:composite}.

\section{Discussion}

We have introduced a simple, yet powerful extension of an important and commonly used model for learning representations and justified its superior performance by its ability to learn a more diverse set of features than the standard denoising autoencoder. Instead of learning a denoising autoencoder with just a single level of noise, we exploit the fact that various levels of noise yield different features, which are more global for large values of $\nu$. Starting the training with a high level of noise enables the algorithm to learn these global features first, which are partially retained when the level of noise is lower and the model is learning more local dependencies. 

\cite{why_does_unsupervised} investigated why unsupervised pretraining helps learning a deep neural network and found that the set of functions learnt by pretrained sigmoid neural networks is very different from the ones that are learnt without unsupervised pretraining. In fact, we have investigated a related question, \emph{why does unsupervised pretraining help unsupervised pretraining?} Or, more precisely, since we are getting a large boost of performance even without supervised fine-tuning, \emph{why does unsupervised pretraining help unsupervised training?} One of their conclusions was that, when using their architecture, unsupervised pretaining puts the optimisation procedure in a basin of attraction of a local minimum that would not otherwise be found. This is very similar to what we observe in our experiments. We often find that a DA trained with a given level of noise $\nu$ can have a lower reconstruction error than ScheDA trained with the final level of noise $\nu$, yet ScheDA is performing better in terms of classification error. The filters at the minima for DA and ScheDA also look very different (cf. \autoref{fig:CIFAR10_filters_low_high}).

This way of training a denosing autoencoder is related to walkback training \citep{generative_DA} in the sense that at the initial stages of training both methods attempt to correctly reconstruct corrupted examples that lie further from the data manifold. It is different though as we do not require the loss to be interpretable as log-likelihood and we do not perform any sampling from the denoising autoencoder.
Additionally, \cite{adaptive_noise} independently tried an idea similar to ScheDA, but they were unable to show consistent improvement over the results of \cite{stacked}.

There is a number of ways this work can be extended. Primarily, ScheDA can be stacked, which would likely improve our results. 
More generally, our results suggest that large improvements can be achieved by combining diverse representations, which we aim to exploit in composite denoising autoencoders.

Finally, we would like to point out that the main observation we make, namely, that it is beneficial for the feature learning algorithm to learn more global features first and then to proceed to learning more local ones, is very general and it is likely that \textit{scheduling} is applicable to other approaches to feature learning. Indeed, in the case of dropout \citep{dropout}, \cite{annealed_dropout} have, independently from our work, explored the use of a schedule to decrease the dropout rate.

\subsubsection*{Acknowledgments}
We thank Amos Storkey, Vittorio Ferrari, Iain Murray, Chris Williams and Ruslan Salakhutdinov for insightful comments on this work.

\subsubsection*{References}

\begingroup
\renewcommand{\section}[2]{}
\bibliography{paper}
\endgroup

\bibliographystyle{plainnat}

\newpage

\section*{Supplementary material}

\subsection*{Sampling the level of noise}

As an alternative to a schedule, which sequentially changes the level of noise, 
we tried to sample a different $\nu$ for each mini-batch. We tried two variants of this idea: sampling $\nu$ uniformly from a continuous interval $[0.1, 0.7]$, and sampling $\nu$ from a discrete distribution over the values in the set $\{0.1, 0.2, \ldots, 0.7\}$. Replicating the setup described in Section \ref{sec:cifar10} for a DA with 2000 hidden units, the first method obtained test error of 44.85\% and the second one obtained the test error 46.83\%. Thus, both have performed much worse than DA (0.3). The result of this experiment provides evidence that training a denoising autoencoder with a sequence of noise levels is important for the success of our method.

\subsection*{Supervised fine-tuning}
For completeness, we also tried training a supervised single-layer neural network using parameters of the encoder as the initialisation of the parameters of the hidden layer of the network. We did that for all models in \autoref{tab:best_CIFAR10}. That is, for each size of the hidden layer, we take the best DA and the best ScheDA trained in an unsupervised manner and perform supervised fine-tuning of their parameters. The learning rate, the same for all parameters, was chosen from the set $\{0.00125, 0.00125 \cdot 2^{-1}, \ldots, 0.00125 \cdot 2^{-4}\}$ and the maximum number of training epochs was 2000 (we computed the validation error after each epoch). We report the test error for the combination of the learning rate and the number of epochs yielding the lowest validation error. The numbers are shown in \autoref{tab:best_finetuned_CIFAR10}. Fine-tuning makes the performance of DA and ScheDA much more similar, but the advantage of ScheDA is consistent and its magnitude grows with the size of the hidden layer.

\begin{table}[h!]
\centering
\caption{Test errors on CIFAR-10 data set for the best DA and ScheDA models trained without supervised fine-tuning and their fine-tuned versions.}
\vspace{5pt}
\begin{tabular}{ c  c | c | c | c | c | c |}         
  \cline{3-4} \cline{6-7}
  & & \multicolumn{2}{ c| }{\textbf{DA}} & & \multicolumn{2}{ c| }{\textbf{ScheDA}} \\
  \cline{1-1} \cline{3-4} \cline{6-7}
  \multicolumn{1}{|c|}{\textbf{hidden  units}} & & \textbf{no fine-tuning} & \textbf{fine-tuning} & & \textbf{no fine-tuning} & \textbf{fine-tuning}\\
  \cline{1-1} \cline{3-4} \cline{6-7}
  \multicolumn{1}{|c|}{1000} & & 45.34\% & 39.55\% & & 43.01\% & 39.44\%\\
  \cline{1-1} \cline{3-4} \cline{6-7}
  \multicolumn{1}{|c|}{2000} & & 41.95\% & 36.85\% & & 40.1\% & 36.22\%\\
  \cline{1-1} \cline{3-4} \cline{6-7}
  \multicolumn{1}{|c|}{5000} & & 38.64\% & 36.47\% & & 36.77\% & 35.7\%\\
  \cline{1-1} \cline{3-4} \cline{6-7}
\end{tabular}
\label{tab:best_finetuned_CIFAR10}
\end{table}

\subsection*{Sentiment classification}
We also evaluate our idea on a data set of product reviews from Amazon \citep{sentiment_data}, adapting the experimental setting used with the CIFAR-10 data set. The version of the data set we are using contains reviews of products from six domains\footnote{books, dvd, electronics, kitchen \& housewares, music, video} corresponding to high-level categories on Amazon.com. The goal is to classify whether a review is positive or negative based on the review text. For computational reasons, we keep only 3000 most popular words in the entire data set, transforming each example into a binary vector indicating presence or absence of a word. We divide the data set into a training set of 10000 labelled examples and 35000 unlabelled examples, a validation set of 10000 labelled examples and a test set of 10000 labelled examples, each of them consisting equal fractions of positive and negative labelled examples. The six domains are mixed among training, validation and test examples. We set the number of hidden units to 2000.

The baseline, logistic regression trained with raw data obtains the test error of 14.79\%, while the best DA (0.6) yields 13.61\% and the best ScheDA (0.7$\to$0.6) yields 13.41\% error. The relative error reduction is smaller than on the image data, which is not surprising since the raw features are here a much stronger baseline and the improvement obtained by the standard DA is relatively smaller too. Smaller relative error reduction can be explained by the fact that the DA performance varies less with the level of noise for this data set. While the test error for the best set of features learnt by DA (0.6) was 13.61\%, the worst, DA (0.1), yielded the error of 13.9\%. This result suggests a simple diagnostic for whether ScheDA is likely to be effective, namely, to check whether the DA validation error is sensitive to the noise level.

\end{document}